\documentclass[a4paper,fleqn]{cas-dc}

\let\printorcid\relax
\usepackage[authoryear]{natbib}
\usepackage[justification=centering]{caption}
\def\tsc#1{\csdef{#1}{\textsc{\lowercase{#1}}\xspace}}
\tsc{WGM}
\tsc{QE}
\tsc{EP}
\tsc{PMS}
\tsc{BEC}
\tsc{DE}

\begin{document}
\let\WriteBookmarks\relax
\def\floatpagepagefraction{1}
\def\textpagefraction{.001}
\shorttitle{HeightFormer}
\shortauthors{Zhan Chen et~al.}

\title [mode = title]{HeightFormer: A Multilevel Interaction and Image-adaptive Classification-regression Network for Monocular Height Estimation with Aerial Images}


\author[1,2,3,4]{Zhan Chen}
\ead{chenzhan21@mails.ucas.ac.cn}

\author[1,2]{Yidan Zhang}
\ead{zhangyidan19@mails.ucas.ac.cn}

\author[1,2,3,4]{Xiyu Qi}
\ead{qixiyu20@mails.ucas.ac.cn}

\author[1,2,3,4]{Yongqiang Mao}
\ead{maoyongqiang19@mails.ucas.ac.cn}
\author[1,2,3,4]{Xin Zhou}
\ead{zhouxin191@mails.ucas.ac.cn}
\author[1,2,3,4]{Lulu Niu}
\ead{niululu17@mails.ucas.ac.cn}
\author[1,2,3,4]{Hui Wu}
\ead{wuhui21@mails.ucas.ac.cn}
\author[1,2,3,4]{Lei Wang\corref{mycorrespondingauthor}}
\cormark[1]
\ead{wanglei002931@aircas.ac.cn}
\author[1,2]{Yunping Ge}
\ead{geyp@aircas.ac.cn}
\address[1]{Aerospace Information Research Institute CAS, No.9 Dengzhuang South Rd, Beijing 100094, CHina}
\address[2]{Key Laboratory of Network Information System Technology (NIST), Aerospace Information Research Institute, Chinese Academy of Sciences, Beijing 100190, China}
\address[3]{University of Chinese Academy of Sciences, No.1 Yanqihu East Rd, Beijing 101408, CHina}
\address[4]{ School of Electronic, Electrical and Communication Engineering, University of Chinese Academy of Sciences, Beijing 100190, China}

\cortext[cor1]{Corresponding author at: Aerospace Information Research Institute CAS, No.9 Dengzhuang South Rd, Beijing 100094, CHina}

\begin{abstract}
Height estimation has long been a pivotal topic within measurement and remote sensing disciplines, proving critical for endeavours such as 3D urban modelling, mixed reality (MR), and autonomous driving. Traditional methods utilise stereo matching or multisensor fusion, both well-established techniques that typically necessitate multiple images from varying perspectives and adjunct sensors like synthetic aperture radar (SAR), leading to substantial deployment costs. Single image height estimation has emerged as an attractive alternative, boasting a larger data source variety and simpler deployment. However, current methods suffer from limitations such as fixed receptive fields, a lack of global information interaction, leading to noticeable instance-level height deviations. The inherent complexity of height prediction can result in a blurry estimation of object edge depth when using mainstream regression methods based on fixed height division.This paper presents a comprehensive solution for monocular height estimation in remote sensing, termed HeightFormer, combining multilevel interactions and image-adaptive classification-regression. It features the Multilevel Interaction Backbone (MIB) and Image-adaptive Classification-regression Height Generator (ICG). MIB supplements the fixed sample grid in CNN of the conventional backbone network with tokens of different interaction ranges. It is complemented by a pixel-, patch-, and feature map-level hierarchical interaction mechanism, designed to relay spatial geometry information across different scales and introducing a global receptive field to enhance the quality of instance-level height estimation. The ICG dynamically generates height partition for each image and reframes the traditional regression task, using a refinement from coarse to fine classification-regression that significantly mitigates the innate ill-posedness issue and drastically improves edge sharpness. Finally, the study conducts experimental validations on the Vaihingen and Potsdam datasets with results demonstrating that our proposed method surpasses existing techniques. The code will be open-sourced at https://github.com/qbxfcz/HeightFormer.

\end{abstract}



\maketitle

\section{Introduction}
\begin{figure}[!ht] 
    \centering
        \includegraphics[width=0.5\textwidth]{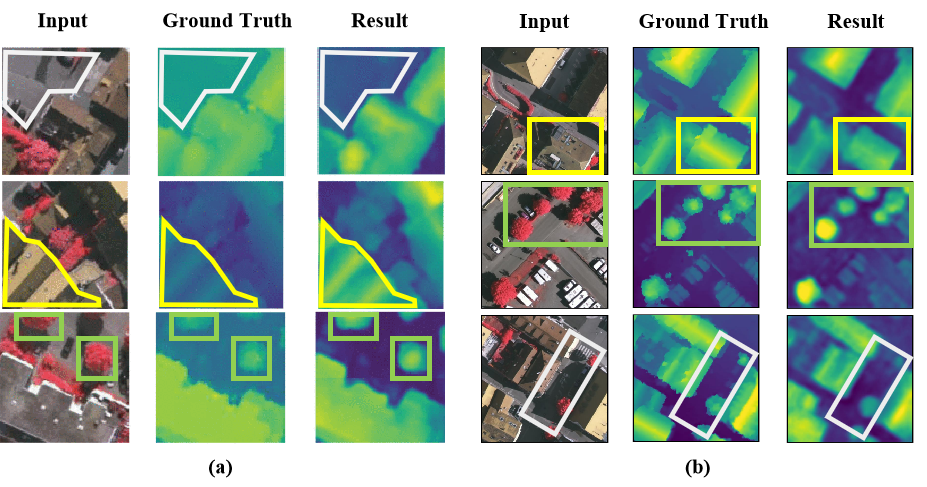} 
    \caption{Typical problems of existing height estimation methods. (a): Instance-level height deviation caused by fixed receptive field. (b): Edge ambiguity caused by fixed height value division(Gray box: road; Yellow box: building; Green box: tree).} 
\label{fig1} 
\end{figure}
With the advancement of high-resolution sensors\\ (\citeauthor{benediktsson2012very}, \citeyear{benediktsson2012very}; \citeauthor{sun2022fair1m}, \citeyear{sun2022fair1m}), the pixel resolution of the visible light band for ground observation has reached the level of decimeters or even centimeters. This has made various downstream applications such as fine-grained urban 3D reconstruction (\citeauthor{zhao2023review}, \citeyear{zhao2023review}), high-precision mapping (\citeauthor{mahabir2018critical}, \citeyear{mahabir2018critical}), and MR scene interaction (\citeauthor{coronado2023integrating}, \citeyear{coronado2023integrating}) incrementally achievable. However, ground height values (Digital Surface Model, DSM), remain at the meter-level (\citeauthor{takaku2021overview}, \citeyear{takaku2021overview}), complicating the execution of various downstream tasks.\\

\begin{figure*}
    \centering 
        \includegraphics[width=0.8\textwidth]{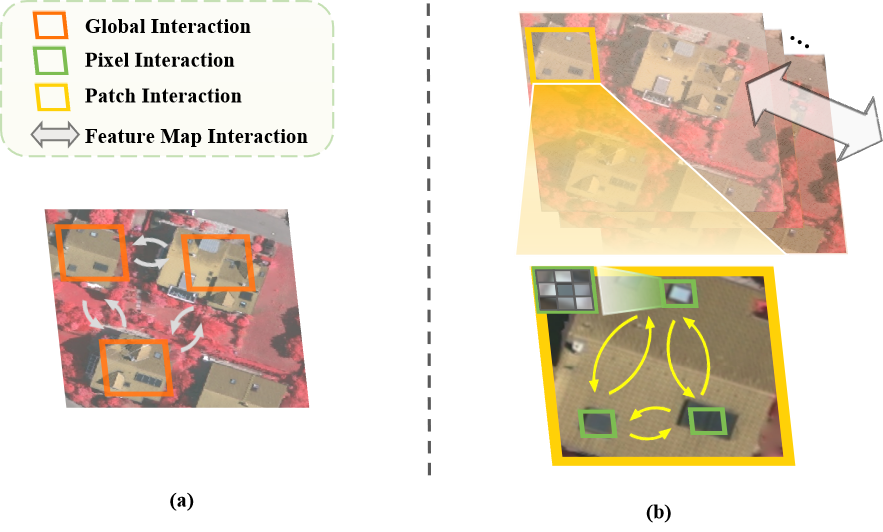}
    \caption{Different interaction mechanism, (a): ViT based interaction mechanism. (b): Our interaction mechanism.} 
\label{fig2} 
\end{figure*}

To improve the accuracy of height estimation, numerous methods have been proposed. For instance, certain studies have expanded research on stereo matching (\citeauthor{nemmaoui2019dsm}, \citeyear{nemmaoui2019dsm}), fitting height information using prior knowledge of viewpoint differences in multiple or multi-view images. Some methods(\citeauthor{zhao2023domain}, \citeyear{zhao2023domain}; \citeauthor{wang2022unetformer}, \citeyear{wang2022unetformer}) further integrate stereo matching and SAR interferometric images, requiring adaptation based on camera parameters, sensor types, and viewpoint differences, and thus have weak transferability. Another approach is to estimate height based on LiDAR generation (\citeauthor{estornell2011analysis}, \citeyear{estornell2011analysis}) , which typically involves using airborne LiDAR sensors to measure distances and subsequently generate height values. However, this method is subject to limitations in terms of sensor platform restrictions and measurement costs.\par
Monocular height estimation methods have low data acquisition cost and easy deployment. In recent years, a series of monocular height estimation methods based on CNNs (\citeauthor{li2020deep}, \citeyear{li2020deep}) have been developed in the field of computer vision. These methods offer good reconstruction quality for details such as edges. However, since monocular height estimation is an inherently complex problem (\citeauthor{kuznietsov2017semi}, \citeyear{kuznietsov2017semi}), it demands the long-range correlation. Given that CNNs have a fixed receptive field, they struggle to conduct whole-image-level information interaction, which leads to common issues such as instance-level height deviation like Fig.\ref{fig1} (a), hindering their widespread application for large area instances of farmland, buildings, etc.\\

\begin{figure}[htbp] 
    \centering 
        \includegraphics[width=0.5\textwidth]{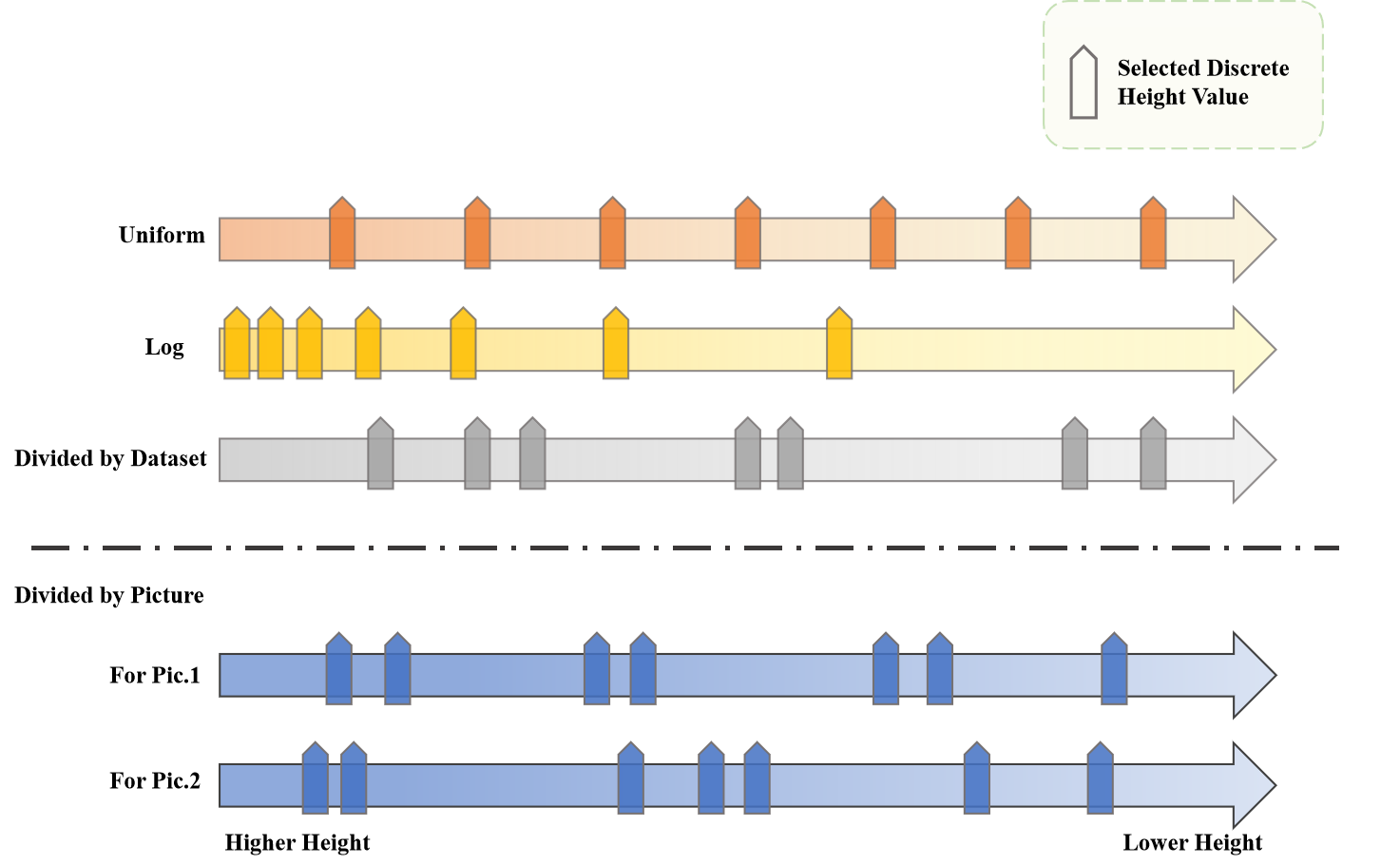} 
    \caption{Different height value generation method, Upper: Fixed height values. Lower: Image-adaptive height values.} 
\label{fig3} 
\end{figure}
But with the advent of Transformers, their attention mechanisms,(\citeauthor{vaswani2017attention}, \citeyear{vaswani2017attention}), which capture long-distance feature dependencies, have significantly improved the inadequate whole-image information interaction common to CNN models. Transformers have been progressively applied in remote sensing tasks such as object detection (\citeauthor{zhao2023domain}, \citeyear{zhao2023domain}) and semantic segmentation (\citeauthor{he2022transformer}, \citeyear{he2022transformer}). However, as shown in Figure \ref{fig2} (a), the transformer based on ViT (\citeauthor{dosovitskiy2020image}, \citeyear{dosovitskiy2020image}) directly interacts over the original resolution of the image. This leads to high modeling complexity. Moreover, the height prediction task requires modeling the height values of all image pixels, further intensifying the difficulty of model convergence. Some methods (\citeauthor{fu2018deep}, \citeyear{fu2018deep}) earlier tried to reduce the difficulty of model convergence by using the classification-regression mechanism, which first predicted the probabilities of different discrete height values and then fitted the final continuous height values. However, as shown in Fig. \ref{fig3}, the methods for setting height values mainly use uniform partitioning, logarithmic partitioning, and partitioning according to the dataset, etc. Due to the large difference in height distribution in different regions of the data set, their edge modeling accuracy is often mediocre like Fig. \ref{fig1} (b). Adabins (\citeauthor{bhat2021adabins}, \citeyear{bhat2021adabins}) adopted image-adaptive strategy earlier, but it only introduced ViT in the decoder stage, the global feature acquisition is still limited, and the quality of edge reconstruction is poor.\par
In this paper, we propose a comprehensive network for height estimation, called HeightFormer, that uses multilevel interaction like Fig. \ref{fig2} (b) and image-adaptive classification-regression. In the encoder part of the network, there are three distinct interaction range feature extraction modules: a convolutional backbone network for pixel-level feature extraction, a local attention backbone network for patch-level feature extraction, and a heterogeneous feature coupling module for global interaction of cross-source features. These modules are designed to bypass the computational complexity of ViT while ensuring that information interaction takes place at various scales.

In the decoder part of HeightFormer, there is an Image-adaptive Classification-regression height Generator (ICG) built on a three-layer multi-head attention based network that is designed to predict the specific number of height values for each image like classification types. In order to obtain the final continuous height values like the regression task. It then multiplies these specific height values with respective corresponding height probabilities obtained through dot product computation to derive the final pixel height values. Compared with fixed height value classification, this method is easier to match the height distribution of different images by generating image-level classification types, and improves the quality of edge reconstruction.\par
Our contributions can be summarized as follows:\par
1.We propose a monocular height estimation method for aerial images, named HeightFormer, which uses multilevel interaction and image-adaptive classification-regression, and achieves robust pixel and instance reconstruction quality.\par
2.The proposed Multilevel Interaction Backbone (MIB) is based on a clever combination of convolution and transformer, resulting in a substantially reduced encoder size while achieving multilevel interaction.\par
3.A new Imgae-adaptive Classification-regression height Generator (ICG) has been proposed, which generates different height value classification types for different images, effectively improving the edge blurring problem associated with fixed height value classification types.\par
4.Experiments are conducted on the Vaihingen and Potsdam datasets, with comparison results indicating that HeightFormer outperforms current remote sensing and  computer vision methods while achieving higher Rel metrics with a smaller number of parameters.\par
The rest of this paper is organized as follows. In Section II, we briefly introduce the related work. Section III explains the details of HeightFormer framework, and extensive experiments are presented in Section IV. Finally, Section V concludes this article.

\section{Related Work}
\noindent\textbf{2.1 Overview}\par
Height estimation is a key component of 3D scene understanding (\citeauthor{wojek2012monocular}, \citeyear{wojek2012monocular} and has long held a significant position in the domains of remote sensing and computer vision. Initial research was largely focused on stereo or multi-view image matching. These methodologies (\citeauthor{goetz2018modeling}, \citeyear{goetz2018modeling}) generally relied on geometric relationships for keypoint matching between two images or more, followed by utilizing triangulation and camera pose data to compute depth information. Recently, the advent of large-scale depth datasets (\citeauthor{geiger2013vision}, \citeyear{geiger2013vision})has led to a shift in research focus. The effort is now centred on estimating distance information from monocular 2D images utilizing supervised learning. Present monocular height estimation approaches can be generally categorized into three types(\citeauthor{li2021geometry}, \citeyear{li2021geometry}; \citeauthor{mou2018im2height}, \citeyear{mou2018im2height}; \citeauthor{yu2021automatic}, \citeyear{yu2021automatic}; \citeauthor{mahdi2020aerial}, \citeyear{mahdi2020aerial}): methodologies based on handcrafted features, methodologies utilizing convolutional neural networks (CNN), and methodologies based on attention mechanisms. In general, because the datasets contain various types of features, it is difficult to extract only manual features to fit the distribution of different datasets, and the effect is generally mediocre.\\\\
\textbf{2.2 Height estimation based on manual features}\par
Conditional Random Fields (CRF) and Markov Random Fields (MRF) are primarily used by researchers to model the local and global structures of images. Since local features alone are inadequate for predicting depth values, (\citeauthor{batra2012learning}, \citeyear{batra2012learning}) simulated the relationships between adjacent regions and used CRF and MRF to model the local and global structures of images. Furthermore, to capture global features beyond local ones, (\citeauthor{saxena2005learning}, \citeyear{saxena2005learning}) computed the features of neighbouring blocks and applied MRF and Laplacian models to estimate the depth of each area. In a separate study, (\citeauthor{saxena2007depth}, \citeyear{saxena2007depth}) introduced the concept of superpixels and implemented them as replacements for pixels during the training process. (\citeauthor{liu2014discrete}, \citeyear{liu2014discrete}) previously formulated the depth estimation issue as a discrete-continuous optimization problem. Here, the discrete component encodes the relationships between adjacent pixels while the continuous part signifies the depth of superpixels. The discrete and continuous variables are interconnected to form a CRF for predicting depth values. (\citeauthor{zhuo2015indoor}, \citeyear{zhuo2015indoor}) introduced a hierarchical approach that amalgamates local depth, intermediate structures, and global structures for depth estimation.\\\\
\textbf{2.3 Height estimation based on CNN}\par
Convolutional Neural Networks (CNNs) have been widely utilized in recent years across various fields of computer vision, including, but not limited to, scene classification, semantic segmentation, and object detection (\citeauthor{zhang2023bridging}, \citeyear{zhang2023bridging}; \citeauthor{zhang2021learning}, \citeyear{zhang2021learning}). Among these tasks, ResNet (\citeauthor{he2016deep}, \citeyear{he2016deep}) is frequently used as the backbone of the models. In a study referred to as IMG2DSM (\citeauthor{ghamisi2018img2dsm}, \citeyear{ghamisi2018img2dsm}), an adversarial loss function was introduced early to enhance the synthesization of Digital Surface Models (DSM), deploying conditional generative adversarial networks to create the transformation from images to DSM elevation. (\citeauthor{zhang2019multi}, \citeyear{zhang2019multi}) improved the ability to learn abstract features of objects at different scales through multi-path fusion networks for multi-scale feature extraction. (\citeauthor{li2020height}, \citeyear{li2020height}) sectioned the height values into intervals with incrementally increasing spacing and reframed the regression problem into an ordinal regression problem, using ordinal loss for network training. Subsequently, they designed a post-processing technique to convert the predicted height maps of each block into seamless height maps. (\citeauthor{carvalho2018regression}, \citeyear{carvalho2018regression}) carried out in-depth research on various loss functions for depth regression, combining encoder-decoder architecture with adversarial loss, and proposed D3Net. (\citeauthor{zhu2016real}, \citeyear{zhu2016real}) aimed to reduce processing time and eliminate fully connected layers before the upsampling process in the visual geometry group. (\citeauthor{kuznietsov2017semi}, \citeyear{kuznietsov2017semi}) enhanced the performance of the network utilizing stereo images with sparse ground truth depths. Its loss function they used harnessed the predicted depth, reference depth, and the differences between the image and the generated distorted image. In conclusion, the convolution based height estimation method achieves the basic fitting of the data set, but it still has obvious deviation in the height prediction at the instance level due to the limitation of fixed receptive field.\\\\
\textbf{2.4 Attention and Transformer in remote sensing}\\
2.4.1 Attention mechanism and Transformer\par
The attention mechanism (\citeauthor{vaswani2017attention}, \citeyear{vaswani2017attention}), an information processing method that simulates the human visual system, enables the assignment of weights to different elements in a learning input sequence. By learning to assign higher weights to essential elements, the attention mechanism permits the model to focus more on critical information, thus enhancing its performance in processing sequential information. In computer vision, the attention mechanism guides the model's focus toward key areas of an image, thereby enhancing model performance. This mechanism can be considered as a simulation of the human perception process of images, that is, understanding the entire image by concentrating on important parts.\par
In recent years, attention mechanisms have been introduced into the field of computer vision due to the exceptional performance of Transformer models in natural language processing (NLP, \citeauthor{vaswani2017attention}, \citeyear{vaswani2017attention}). These models, which are based on a self-attention mechanism (\citeauthor{shaw2018self}, \citeyear{shaw2018self}), establish global dependencies in the input sequence, thus enabling them to better handle sequential information. However, the format of input sequences in computer vision differs from that in NLP, encompassing various formats such as vectors, single-channel feature maps, multi-channel feature maps, and feature maps derived from differing sources. As a result, attention mechanisms such as spatial attention (which learns attention between different regions of feature maps, \citeauthor{jaderberg2015spatial}, \citeyear{jaderberg2015spatial}), local attention (which limits the computational span of spatial attention, \citeauthor{luong2015effective}, \citeyear{luong2015effective}), cross-channel attention (which calculates attention between different channels of feature maps,  \citeauthor{hu2018squeeze}, \citeyear{hu2018squeeze}), and cross-modal attention (which calculates attention between feature maps or feature vectors from different sources, \citeauthor{huang2019ccnet}, \citeyear{huang2019ccnet}; \citeauthor{vaswani2017attention}, \citeyear{vaswani2017attention}), need to be adapted for visual sequence modeling.\par
Concurrently, various visual Transformer methods have been proposed, including the Vision Transformer (ViT, \citeauthor{dosovitskiy2020image}, \citeyear{dosovitskiy2020image}), which partitions an image into blocks and transforms these blocks into vectors to compute attention between them. The Swin Transformer (\citeauthor{liu2021swin}, \citeyear{liu2021swin}), significantly reduces the computational burden by setting three different attention computation scales and optimizes the model's ability to perform local modeling across different visual tasks.\\\\
2.4.2 Transformers applied in remote sensing\par
Since 2023, numerous remote sensing tasks have progressively introduced or optimized Transformer networks.  (\citeauthor{yang2023center}, \citeyear{yang2023center}) were among the pioneers to utilize an optimized Vision Transformer (ViT) network for hyperspectral image classification, adjusting the sampling method of ViT to enhance local modeling. In the object detection domain,  (\citeauthor{zhao2023domain}, \citeyear{zhao2023domain}) integrated two additional classification tokens for Synthetic Aperture Radar (SAR) images to improve the accuracy of ViT-based detection. (\citeauthor{chen2023large}, \citeyear{chen2023large}) introduced the MPViT network, bringing together scene classification, super-resolution, and instance segmentation to boost the quantity and area of building extraction substantially. In relation to the supervised paradigm, (\citeauthor{he2023ast}, \citeyear{he2023ast}) were early contributors who integrated a pyramid structure into ViT to broaden the self-supervised learning applications in optical remote sensing image interpretation. To summarize, in the field of remote sensing, existing methods often enhance the ViT structure with local modeling capabilities to accommodate the multi-scale nature of remote sensing images. Our approach combines feature extraction modules at different interaction scales to heighten the robustness of our model.

\section{Methodology}
\noindent\textbf{3.1 Overview}\par

\begin{figure*}
    \centering
        \includegraphics[width=1\textwidth]{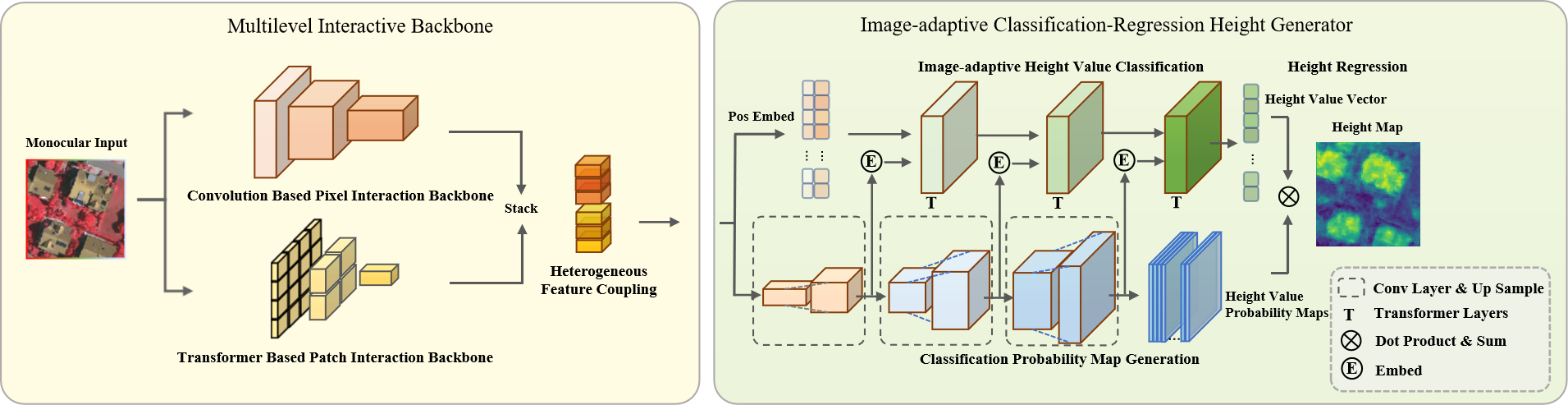} 
    \caption{Architecture of HeightFormer, consisting of Multilevel Interactive Backbone (MIB) and Image-adaptive Classification-Regression Height Generator (ICG).} 
\label{fig4} 
\end{figure*}
The architecture of the HeightFormer, as depicted in Fig.\ref{fig4}, comprises primarily of two components: the Multilevel Interactive Backbone (MIB) and the Image-adaptive Classification-Regression Height Generator (ICG). As an encoder, the MIB consists of three tiered feature extraction modules: the Pixel Interactive Backbone, Patch Interactive Backbone, and the Heterogeneous Feature Coupling Module. Each module progressively expands its feature interaction range to perform feature extraction at various scales. The decoder section incorporates an image-adaptive classification-regression module equipped with a multi-head attention based transformer branch. This allows it to predict the discrete height values of an individual image. To tackle the multi-scale nature of aviation images, HeightFormer uses multi-scale convolutional branches to gradually reconstruct high-resolution height probability maps from low-resolution feature maps. The definitive height value is derived from the product of probabilities at differing height levels for a single pixel.\\\\
\textbf{3.2 Multilevel Interactive Backbone}\par

\begin{figure*} 
    \centering 
        \includegraphics[width=1\textwidth]{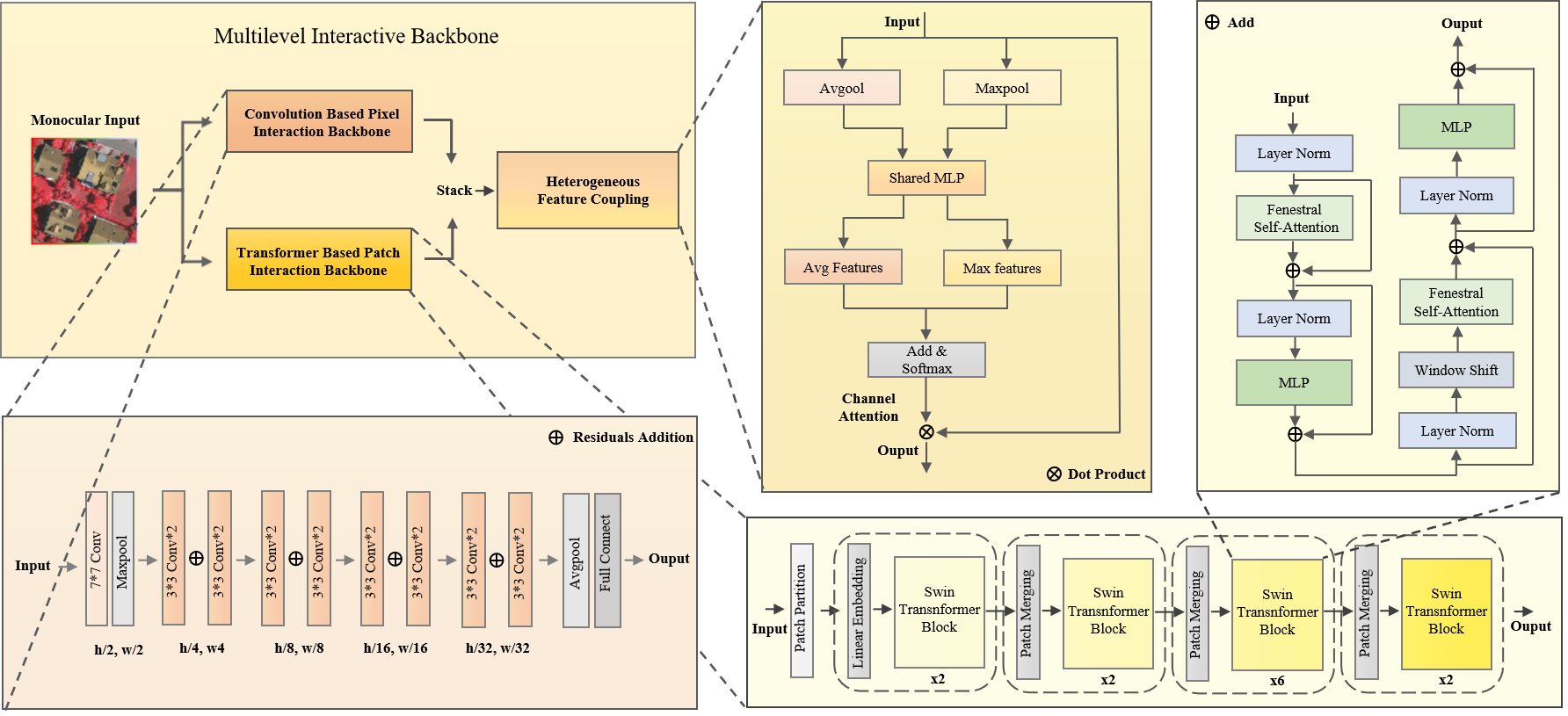} 
    \caption{Architecture of Multilevel Interactive Backbone, mainly consisting of Convolution Based Pixel Interaction Backbone, Transformer Based Patch Interaction Backbone and Heterogeneous 
Feature Coupling.} 
\label{fig5} 
\end{figure*}
The encoder incorporates two backbone networks and a coupling module. The dual backbone networks comprise a convolutional branch, aimed at extracting pixel-level features, and a local attention branch designated for obtaining patch-level features. The Heterogeneous Feature Coupling module employs cross-channel attention to facilitate larger-scale feature interactions.\\\\
3.2.1 Convolution Based Pixel Interaction Backbone\par
As shown in Fig. \ref{fig5} Convolution Based Pixel Interaction Backbone, 
to limit the total number of model parameters as well as to ensure the extraction of primary pixel-level features through the convolution branch, a ResNet18 network is used as the anchor for pixel-level feature extraction. The architecture of the ResNet18 network encompasses five stages of convolution modules which sequentially transform the feature map into an N$\times$7$\times$7 format. The final transformation to the prescribed dimension (16N, N is the number of height categories) is facilitated through a fully connected layer. The entire process comprises the Conv(7, 7), Maxpool, Conv(3, 3), Avgpool, residual calculation,  and Full Connect of the output layer.\\\\
3.2.2 Transformer Based Patch Interaction Backbone\par
In the Transformer
based patch interaction backbone, unlike Vit which uses fixed patch divisions, we utilize the scale pyramid and feature interaction window shift characteristics of Swin Transformer to control the feature interaction range between adjacent pixel interaction and global interaction. We use the Swin Transformer Tiny as the backbone network. The Swin Transformer sets four stages of downsampling and attention scope. By separately computing the Fenestral Self-Attention and shift-after Fenestral Self-Attention in each Swin Transformer Block, the range of feature interaction is gradually enlarged. We set the output dimension to 16N.The computational complexities of the standard Multi-head Self-Attention ($MSA$) and Fenestral Self-Attention ($FSA$) are respectively:\par
\begin{equation}
  \Omega (MSA)=4hwC^{2} +2(hw)^{2} C
\end{equation}
\begin{equation}
  \Omega (FSA)=4hwC^{2} +2M^{2} hwC
\end{equation}\par
Here, $h,w,C,M$ represent the image height, image width, image channel, and the division window, respectively. \par
Fenestral Self-Attention reduces the complexity related to image height and width, and substitutes it with window division, which is more suitable for aviation or remote sensing images with a wide range of scales and resolutions.\par
In a Swin Transformer Block, given an input denoted by $z ^{l-1}$, the corresponding output, represented by $z ^{l+1}$, can be calculated as follows:\par
\begin{equation}
  \hat{z} ^l=FSA(LN(z ^{l-1}))+z ^{l-1}
\end{equation}
\begin{equation}
z^l=MLP(LN(\hat{z}^l))+\hat{z}^l
\end{equation}
\begin{equation}
  \hat{z} ^{l+1}=FSA(WS(LN(z ^{l})))+z ^{l}
\end{equation}
\begin{equation}
z^{l+1}=MLP(LN(\hat{z}^{l+1}))+\hat{z}^{l+1}
\end{equation}
\\\\
In this context, "LN" stands for Layer Normalization, "MLP" denotes a Multi-layer Perceptron and "WS" represents Window Shift.\\\\
3.2.3 Heterogeneous feature coupling\par
The feature maps extracted by convolution and transformer have the multichannel (respectively 256N) and pony-size (usually 1/16 of the original width/height) characteristics. Heterogeneous feature coupling calculates the weights and fuses the two feature maps in the channel dimension. Like SENet (\citeauthor{hu2018squeeze}, \citeyear{hu2018squeeze}), assuming the stacked input feature is $X$, the features after max pooling and average pooling are subjected to shared MLP operations indicated as $\hat{X_1}$ and $\hat{X_2}$. The sum of the two features is normalized to obtain the attention weight $W_{Attention}$, and the final output $Y$ (256N) is obtained by the dot product of the attention weight $W_{Attention}$ and the original input $X$. Specifically, it can be represented as:\par
\begin{equation}
\hat{X_1},\hat{X_2}=MLP(Avgpool(X),Maxpool(X))
\end{equation}
\begin{equation}
    W_{Attention} =SoftMax(\hat{X_1}+\hat{X_2})
\end{equation}
\begin{equation}
    Y=W_{Attention}\otimes X 
\end{equation}
\\\\
Heterogeneous feature coupling achieves feature redistribution of the two feature maps by calculating attention weights, capturing key information at different feature granularities.\\\\
\textbf{3.3 Image-adaptive Classification-Regression Height Generator}
\begin{figure*}
    \centering 
        \includegraphics[width=0.9\textwidth]{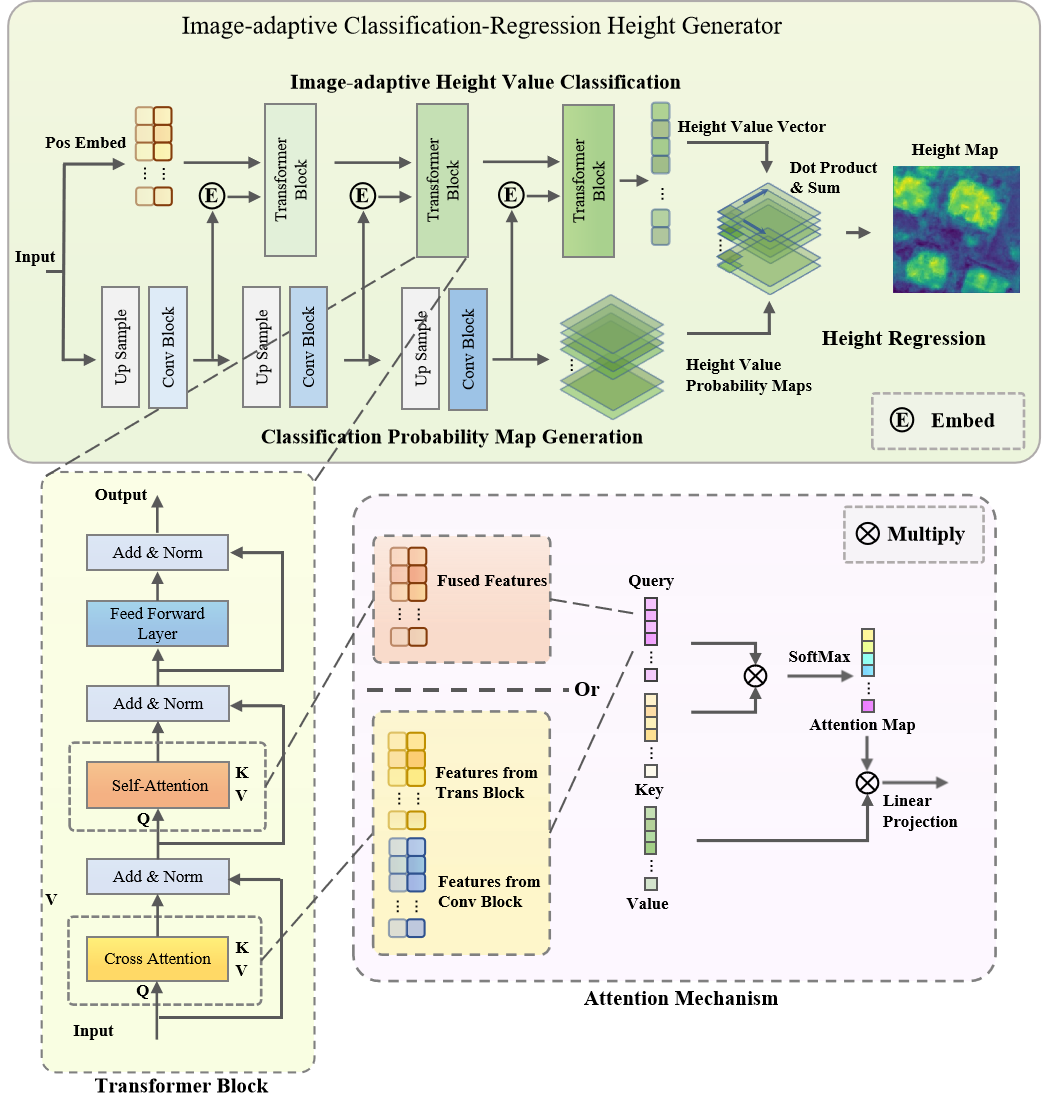} 
    \caption{Architecture of Image-adaptive Classification-Regression Height Generator, consisting of Image-adaptive Height Value Classification and Classification Probability Map Generation. } 
\label{fig6} 
\end{figure*}\par

Fig.\ref{fig6} depicts the components of the Image-adaptive Classification-Regression Height Generator, which consists of Image-adaptive Height Value Classification, Classification Probability Map Generation, and Height Regression. During the encoder stage, we continually refine the feature map size for effective feature fusion. Correspondingly, in the decoder stage, receiving input $Y$ (H/16, W/16, 256N) from the encoder, we enhance multi-scale feature maps to accomplish fine-grained modeling. We employ a Transformer Block and a Convolution Block, at three distinct scales: (H/16, W/16, 256N), (H/4, W/4, 16N), and (H, W, N), to categorize height values of the input image and produce height probability maps. Subsequently, we obtain the height value regression by calculating the product of height values and probability maps.\\\\
3.3.1 Image-adaptive Height Value Classification\par
In the Image-adaptive Height Value Classification section, we obtain a 1$\times$N height vector(denoted as $H$), where each height vector value corresponds to a classification type, with N representing the number of classification settings determined by the experiment. To reduce the randomness of generated height values and increase the affinity between the two branches, cross-attention is incorporated into the initial layer of the Transformer Block to further interact with the probability graph branch (denoted as $P$). This means that the current Transformer Block layer accepts queries output from the previous layer and outputs embedded from the Conv Block on the same layer. The Transformer Block also includes addition\&normalization, self-attention, and feed-forward layer operations. If the current Transformer Block layer is the l layer, then the output of the previous Transformer Block is represented as $H^{l-1}$, and the input to the Conv Block is represented as $P^{l}$; the operation process can be described as follows:\par
\begin{equation}
    \hat H^{l}=LN(CA(H^{l-1},Embed(P^{l}))+H^{l-1})
\end{equation}
\begin{equation}
    \tilde{H}^{l}=LN(MSA(\hat H^{l})+\hat H^{l})
\end{equation}
\begin{equation}
H^{l}=LN(FFL(\tilde H^{l}))+\tilde H^{l})
\end{equation}\par
The symbols $CA, MSA, LN$, and $FFL$ respectively denote Cross Attention, Multi-Head Self Attention, Layer Normalization, and Feed Forward Layer operations.\\\\
3.3.2 Classification Probability Map Generation\par
For Classification Probability Map Generation, it receives a high channel feature map (256N) and outputs a height probability map of size (H, W, N). To adapt to the rich scale characteristics of aerial image targets, we have designed a three-layer convolution pyramid structure with up sampling to continuously rebuild the detail features. The Convolution Block we designed includes UpSample, $Conv_{3\times3}$, $Conv_{1\times1}$, and $ReLu$; the output sizes of the three layers are (H/16, W/16, 16N), (H/4, W/4, 4N), and (H, W, N). If the current layer is denoted as $l$, and the input from the previous Convolution Block is denoted as $P^{l-1}$, then the calculation process can be represented as:
\begin{equation}
   \hat P^{l}=ReLu(Conv_{3\times 3}(UpSample(P^{l-1})))
\end{equation}
\begin{equation}
    P^{l}=LN(Conv_{1\times1}(\hat P^{l}))
\end{equation}\\
3.3.3 Height Regression\par
During the execution of the Height Regression section, we initially obtain the Height Vector ($H$) and the Height Value Probability Map ($P$) separately. Subsequently, employing SoftMax, we adjust both of them to fit the 0-1 range, and through the dot product operation, we derive a height map of dimensions (H, W, 1). Ultimately, the height values are linearly scaled to comply with the height range of the dataset. The minimum and maximum height values of the dataset are respectively represented by $h_{min}$ and $h_{max}$. The specific process unfolds as follows:\par
\begin{equation}
    \hat H=SoftMax(H) 
\end{equation}
\begin{equation}
    \hat P=SoftMax(P)
\end{equation}
\begin{equation}
    \hat {Result}=\sum_{i=1} ^{N} (\hat  H_{i} \times \hat P_{i})
\end{equation}
\begin{equation}
    Result=h_{min}+(h_{max} -h_{min}) \times \hat{Result}
\end{equation}\\\\
\textbf{3.4 Loss function}\par
We adopt the Sigmoid Cross-Entropy Loss Function, akin to other methodologies (\citeauthor{li2022binsformer},  \citeyear{li2022binsformer}), that can be precise in its representation:\par
\begin{equation}
    g_i=\log_{}{\tilde{h}_i} -\log_{}{{h}_i}
\end{equation}
\begin{equation}
    Loss=\alpha \sqrt{\frac{1}{T}\sum_{i}g_{i}^2- \frac{\lambda }{T^2}(\sum_{i} g_{i})^2}
\end{equation}\par
The symbols $\tilde{h}_i$, ${h}_i$, and $T$ denote the estimated height, the actual height, and the number of valid pixels respectively. Like adabins (\citeauthor{bhat2021adabins}, \citeyear{bhat2021adabins}), the parameters $\lambda$ and $\alpha$ are assigned the values 0.85 and 10 correspondingly.
\section{Experiment}
\noindent\textbf{4.1 Datasets}\par
We utilize the Vaihingen and Potsdam datasets from ISPRS to ecalute the effectiveness of our proposed method, following the official partition of training and testing sets. The Vaihingen dataset comprises 33 images, each approximately 2500$\times$2500 pixels in size. With a pixel resolution of 0.09 meters, these images span a height range from 240.70033 to 360.0037 meters. As is common with other methodologies, we linearly normalize the heights to fall within the 0-1 range. Of the total images, 26 (80\%) serve as the training set while 7 (20\%) are designated for testing. Given the extensive size of the images, it is standard practice to crop them into smaller 512x512 images for both training and testing applications.\par
The Potsdam dataset includes 38 images, each measuring 6000x6000 pixels. These images, with a pixel resolution of 0.05 meters, have a height range from -17.355 to 106.171 meters. The dataset is divided into 30 (80\%) images for training and 18 (20\%) for testing. As with the Vaihingen dataset, we normalize the heights and crop the images in a consistent manner.\\\\
\textbf{4.2 Metrics}\par
The assessment metrics employed encompass prevailing indicators like Rel, RMSE(log), and threshold values $\delta 1$, $\delta 2$, $\delta 3$. Rel emphasizes the measurement of average errors, while RMSE(log) displays higher sensitivity to outliers possessing substantial errors. Additionally,  $\delta 1$, $\delta 2$, $\delta 3$ represent Threshold Accuracy Metrics, gauging the proportion of pixels maintaining error control within a designated range and prioritizing overall error stability. These metrics are explicitly illustrated as follows:\par
\begin{equation}
    Rel=\frac{1}{n} \sum \left | \frac{h_{pred}-h_{gt}}{h_{gt}}  \right | 
\end{equation}
\begin{equation}
    RMSE(log)=\sqrt{\frac{1}{n} \sum(h_{pred}-h_{gt})^2 } 
\end{equation}
\begin{equation}
    \delta _i=Max(\frac{h_{pred}}{h_{gt}},\frac{h_{gt}}{h_{pred}})<1.25^i
\end{equation}\par
Given $h_{pred}$, $h_{gt}$, and n, which represent the predicted height map, the ground truth height map, and the number of pixels within the height map, respectively. \\\\
\textbf{4.3 Experimental settings}\\
4.3.1 Hardware platform and libraries\par
The training component was executed on 4 RTX 3090 GPUs, while a single RTX 3090 GPU was utilized for testing and comparison. Across these tasks, we employed the Ubuntu 22.04 system, nvidia-driver-525-server driver, CUDA 11.1 computing library, and the PyTorch 1.8.1 deep learning framework. Multi-threading mechanism like MMSegmentation handled the distributed computing across the 4 RTX 3090 GPUs. For the purpose of comparing parameter quantities, we did not make use of MMCV's dynamic quantization capability.\\\\
4.3.2 Training\par
We train 24 epochs like binsformer (\citeauthor{li2022binsformer}, \citeyear{li2022binsformer}) and depthformer (\citeauthor{li2022depthformer}, \citeyear{li2022depthformer}) on both datasets running a batch size of 2 per GPU and initializing the learning rate at $1\times10^{-5}$. Utilizing the AdamW optimizer, we adjusted the learning rate throughout the training process. To circumvent early training instability, a warm-up period accounting for an eighth of the total training process was integrated during the initial phase.\\\\
4.3.3 Data augmentation\par
We employed MMCV-provided data augmentation methods during the training phase. These include random image cropping to 448 $\times$ 448 dimensions, image rotation with a probability of 0.5 and a degree of 2.5, and both photometric and chromatic augmentations with a probability of 0.5, gamma range of 0.9 to 1.1, brightness range of 0.75 to 1.25, and color range of 0.9 to 1.1. Conversely, during the testing phase, we refrained from utilizing any data augmentation.
\section{Results}
\noindent\textbf{5.1 Quantitative and qualitative analysis on Vaihingen}\par
\begin{table*}[width=1.9\linewidth,cols=7,pos=!htb]
\caption{Method comparison on the Vaihingen dataset}\label{tbl1}
\begin{tabular*}{\tblwidth}{@{} LLLLLLL@{} }
\toprule
Method & Ref & Rel$\downarrow$ & RMSE(log)$\downarrow$ & $\delta$1$\uparrow$ & $\delta$2$\uparrow$ & $\delta$3$\uparrow$ \\
\midrule
D3Net (\citeauthor{carvalho2018regression}, \citeyear{carvalho2018regression}) & ICIP 2018 & 2.016 & - & - & - & -\\
Amirkolaee et al.  (\citeauthor{amirkolaee2019height}, \citeyear{amirkolaee2019height}) & ISPRS 2019 & 1.163 & 0.334 & 0.330 & 0.572 & 0.741\\
PSDNet (\citeauthor{zhou2020pattern}, \citeyear{zhou2020pattern}) & CVPR 2020 & 0.363 & 0.171 & 0.447 & 0.745 & 0.906\\
Li et al. (\citeauthor{li2020height}, \citeyear{li2020height}) & GRSL 2020 & 0.314 & 0.155 & 0.451 & 0.817 & 0.939\\
WMD (\citeauthor{ramamonjisoa2021single}, \citeyear{ramamonjisoa2021single})& CVPR 2021 & 0.272 & - & 0.543 & 0.798 & 0.916\\
LeReS (\citeauthor{yin2021learning}, \citeyear{yin2021learning})& CVPR 2021 & 0.260 & - & 0.554 & 0.800 & 0.932\\
ASSEH (\citeauthor{liu2022associatively}, \citeyear{liu2022associatively})& TGRS 2022 & 0.237 & 0.120 & 0.595 & 0.860 & 0.971\\
DepthsFormer (\citeauthor{li2022depthformer}, \citeyear{li2022depthformer})& CVPR 2022 & 0.212 &0.080 & 0.716 & 0.927 & 0.967\\
BinsFormer (\citeauthor{li2022binsformer}, \citeyear{li2022binsformer})& CVPR 2022 & 0.203 & 0.076 & 0.745 & 0.931 & \textbf{0.975}\\
SFFDE (\citeauthor{mao2023elevation}, \citeyear{mao2023elevation})& TGRS 2023 & 0.222 & 0.084 & 0.595 & 0.897 & 0.970\\
HeightFormer & - &\textbf{0.185} &\textbf{0.074} &\textbf{0.756} & \textbf{0.941} & 0.973\\
\bottomrule
\end{tabular*}
\end{table*}

\begin{figure*}
    \centering 
        \includegraphics[width=0.7\textwidth]{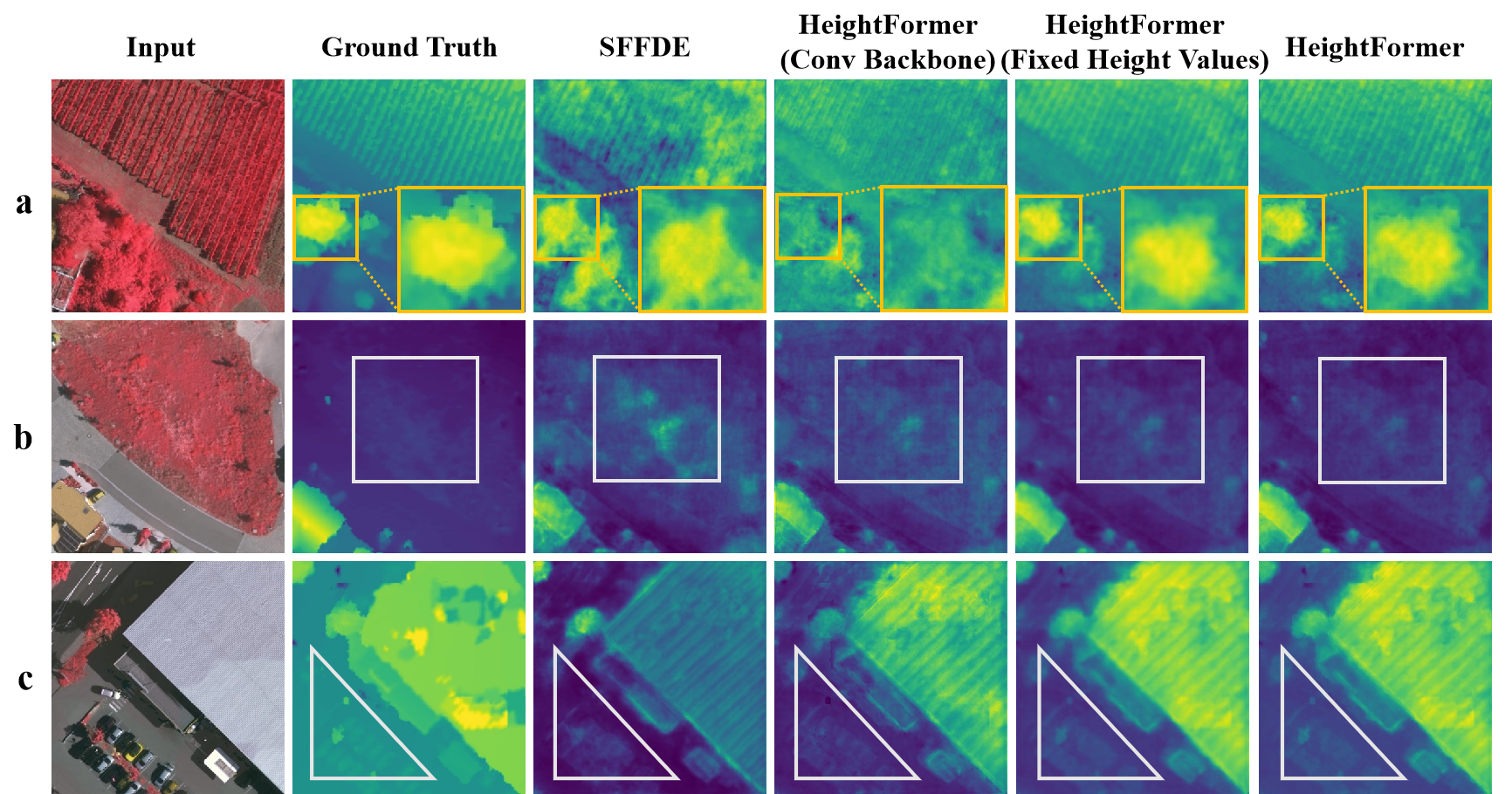} 
    \caption{Visualization Results of Vaihingen (Yellow box: edge; Gray box: instance).} 
\label{fig7} 
\end{figure*}
For the evaluation utilizing Vaihingen dataset, our proposed method, HeightFormer, was juxtaposed with existing models such as D3Net (\citeauthor{carvalho2018regression}, \citeyear{carvalho2018regression}), Amirkolaee et al. (\citeauthor{amirkolaee2019height}, \citeyear{amirkolaee2019height}), PSDNet (\citeauthor{zhou2020pattern}, \citeyear{zhou2020pattern}), Li et al.(\citeauthor{li2020height}, \citeyear{li2020height}), WMD (\citeauthor{ramamonjisoa2021single}, \citeyear{ramamonjisoa2021single}), LeReS (\citeauthor{yin2021learning}, \citeyear{yin2021learning}),  ASSEH (\citeauthor{liu2022associatively}, \citeyear{liu2022associatively}), Depthformer (\citeauthor{li2022depthformer}, \citeyear{li2022depthformer}), Binsformer (\citeauthor{li2022binsformer}, \citeyear{li2022binsformer}), and SFFDE (\citeauthor{mao2023elevation}, \citeyear{mao2023elevation}) at Table \ref{tbl1}. Rel and RMSE(log) primarily quantify the average error in pixel prediction. HeightFormer, which integrates interaction ranges across various scales, achieves state-of-the-art performance. Metrics $\delta 1$,$\delta 2$, and $\delta 3$ determine the percentage of pixels with prediction errors within specific boundaries. The HeightFormer model, being compact, exhibits marginally larger height prediction errors for certain pixels, indicating that robustness warrants further enhancement. In terms of  $\delta 3$, it slightly underperforms BinsFormer (0.973 < 0.975). A visual comparison between SFFDE and HeightFormer outcomes is also presented as Fig. \ref{fig7}. Owing to its capacity to engage with global information, HeightFormer displays improved accuracy in height prediction for extensive scenes. As depicted in Fig. \ref{fig7} (c), this mitigates the instance-level bias issue inherent in convolutional networks.\\\\
\noindent\textbf{5.2 Ablation study on Vaihingen}\par
In the ablation study section, we further tested the effectiveness of the Multilevel Interactive Backbone (MIB) and the Image-adaptive Classification-regression Height Generator (ICG) on Vaihingen.\\
5.2.1 Ablation of Multilevel Interaction Backbone (MIB)\par
\begin{table*}[width=1.9\linewidth,cols=8,pos=htbp]
\caption{Ablation study with different modules of MIB on the Vaihingen dataset}\label{tbl2}
\begin{tabular*}{\tblwidth}{@{} LLLLLLLL@{} }
\toprule
Pixel- & Patch- & Heterogeneous feature coupling & Rel$\downarrow$ & RMSE(log)$\downarrow$ & $\delta$1$\uparrow$ & $\delta$2$\uparrow$ & $\delta$3$\uparrow$ \\
\midrule
$\surd$ & - & - & 0.281 & 0.113 & 0.564 & 0.794 & 0.947\\
- & $\surd$ & - & 0.203 & 0.077 & 0.624 & 0.895 & 0.959\\
$\surd$ & $\surd$ & $\surd$ &\textbf{0.185} & \textbf{0.074} & \textbf{0.756} & \textbf{0.941} & \textbf{0.973}\\
\bottomrule
\end{tabular*}
\end{table*}\par
Table \ref{tbl2} delineates the test outcomes derived from various interactional-level modules within the Multilevel Interactive Backbone (MIB). Evidenced by Line 1 of \ref{tbl2}, the exclusive utilization of a pure convolution backbone network yields subpar performance metrics, while Column 4 in Fig. \ref{fig7} further demonstrates the arduousness of reconstructing comprehensive instances due to insufficient global information exchange. Prevailing state-of-the-art (SOTA) models, such as Binsformer (\citeauthor{li2022binsformer}, \citeyear{li2022binsformer}), are constructed on the foundation of Transformer Backbone, and correspondingly, Line 2 of \ref{tbl2} exhibits analogous performance benchmarks within our model. Ultimately, the HeightFormer effectively orchestrates disparate interaction modules, culminating in superior performance indicators and graphical representations.\\
5.2.2 Ablation of Image-adaptive Classification-Regression Height Generator (ICG)
\begin{table*}[width=1.9\linewidth,cols=6,pos=!h]
\caption{Ablation study with different height value generation strategies of ICG on the Vaihingen dataset}\label{tbl3}
\begin{tabular*}{\tblwidth}{@{} LLLLLLL@{} }
\toprule
Type & N (num of height) & Rel$\downarrow$ & RMSE(log)$\downarrow$ & $\delta$1$\uparrow$ & $\delta$2$\uparrow$ & $\delta$3$\uparrow$ \\
\midrule
\multirow{6}{*}{Fixed} 
& 8 & 0.402 & 0.179 & 0.463 & 0.725 & 0.845\\
& 16 & 0.356 & 0.156 & 0.502 & 0.747 & 0.859 \\
& 32 &  0.314 & 0.129 & 0.581 & 0.813 &
0.862 \\
& 64 & 0.288 & 0.118 & 0.619 & 0.846 &
0.877 \\
& 128 & 0.263 & 0.114 & 0.653 & 0.836 &
0.912 \\
& 256 & 0.267 & 0.118 & 0.639 & 0.826 &
0.920 \\
\midrule
\multirow{6}{*}{Image-adaptive} 
& 8 & 0.341 & 0.135 & 0.458 & 0.742 & 0.903\\
& 16 & 0.307 & 0.119 & 0.519 & 0.795 & 0.938 \\
& 32 &  0.203 & 0.076 & 0.714 & 0.935 & 0.965 \\
& 64 & \textbf{0.185} & \textbf{0.074} & \textbf{0.756} & \textbf{0.941} & \textbf{0.973} \\
& 128 & 0.191 & 0.075 & 0.737 & 0.921 & 0.967 \\
& 256 & 0.224 & 0.084 & 0.673 & 0.901 & 0.959 \\
\bottomrule
\end{tabular*}
\end{table*}

\begin{figure}[!h] 
\includegraphics[width=0.5\textwidth]{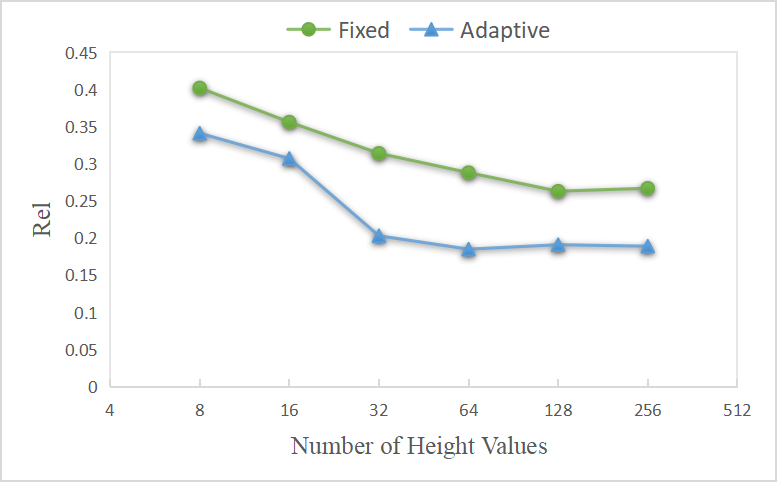} 
\caption{Rel Comparison between Fixed Height Values and Image-adaptive Height Values on Vaihingen.} 
\label{fig8}
\end{figure}\par
In the context of ICG, we assessed the efficacy of the image-adaptive height value approach and investigated its influence on model performance in relation to the various height value number settings. Fig. \ref{fig8} exemplifies this through the Rel indicator, demonstrating that the adaptive height strategy typically yields enhanced Rel indicator outcomes across distinct height value configurations, with other indicator performances detailed in Table \ref{tbl3}. With regard to visualization results, Columns 5 and 6 of Fig. \ref{fig7} reveal that, owing to an improved fit in height distribution for the corresponding input image, the adaptive height strategy exhibits discernible enhancements in edge detail restoration (yellow box) and noise suppression during instance reconstruction (gray box). In reference to the quantity of the height value (N), the Rel index for both methodologies experiences an initial decline, followed by stabilization as the count of height values escalates. Generally, exceedingly minimal height value settings resemble low-type classification tasks, thereby constraining modeling precision. Conversely, a profusion of height value settings approximates intricate regression endeavors, consequently amplifying the model's complexity and impeding convergence. Ultimately, the HeightFormer attains peak efficacy at a proximal estimate of 64 height value settings. Owing to the constrained model fitting capacity inherent in fixed height settings, optimal performance is attained at an expanded height value configuration (approximately 128).\\\\
\textbf{5.3 Method comparison of the computational power consumption}\par
\begin{table*}[width=1.9\linewidth,cols=4,pos=!h]
\caption{Method comparison of model size and inference speed}\label{tbl4}
\begin{tabular*}{\tblwidth}{@{} LLLLLLL@{} }
\toprule
Method & Ref & Parameters & FPS\\
\midrule
Li et al. (\citeauthor{li2020height}, \citeyear{li2020height}) & GRSL 2020 & - & 8.7 \\
DepthsFormer (\citeauthor{li2022depthformer}, \citeyear{li2022depthformer}) & CVPR 2022 & 273M & 8.2 \\
BinsFormer (\citeauthor{li2022binsformer}, \citeyear{li2022binsformer})& CVPR 2022 & 254M & 8.0 \\
SFFDE (\citeauthor{mao2023elevation}, \citeyear{mao2023elevation})& TGRS 2023 & >60M & 8.7 \\
HeightFormer & - & \textbf{46M} & \textbf{10.8} \\
\bottomrule
\end{tabular*}
\end{table*}
In Table \ref{tbl4}, the parameter count and inference speed (Frames Per Second, FPS) on a singular RTX 3090 for the most recent models are extra presented. Leveraging a lightweight backbone network and a multi-scale reconstruction structure, HeightFormer outperforms with respect to both parameter quantity and end-point deployment requirements.\\\\
\noindent\textbf{5.4 Quantitative and qualitative analysis on Potsdam}\par
\begin{table*}[width=1.9\linewidth,cols=7,pos=!h]
\caption{Method comparison on the Potsdam dataset}\label{tbl5}
\begin{tabular*}{\tblwidth}{@{} LLLLLLL@{} }
\toprule
Method & Ref & Rel$\downarrow$ & RMSE(log)$\downarrow$ & $\delta$1$\uparrow$ & $\delta$2$\uparrow$ & $\delta$3$\uparrow$ \\
\midrule
Amirkolaee et al. (\citeauthor{amirkolaee2019height}, \citeyear{amirkolaee2019height})& ISPRS 2019 & 0.571 & 0.259 & 0.342 & 0.601 & 0.782\\
BAMTL (\citeauthor{9288901}, \citeyear{9288901})& J-STARS 2020 & 0.291 & - & 0.685 & 0.819 & 0.897\\
DepthsFormer (\citeauthor{li2022depthformer}, \citeyear{li2022depthformer})& CVPR 2022 & 0.123 &0.050 & 0.871 & 0.981 & 0.997\\
BinsFormer (\citeauthor{li2022binsformer}, \citeyear{li2022binsformer})& CVPR 2022 & 0.117 & 0.049 & 0.876 & 0.989 & \textbf{0.999}\\
HeightFormer & - & \textbf{0.104} & \textbf{0.043} & \textbf{0.893} & \textbf{0.987} & 0.997\\
\bottomrule
\end{tabular*}
\end{table*}

\begin{figure}[!h] 
\centering 
\includegraphics[width=0.4\textwidth]{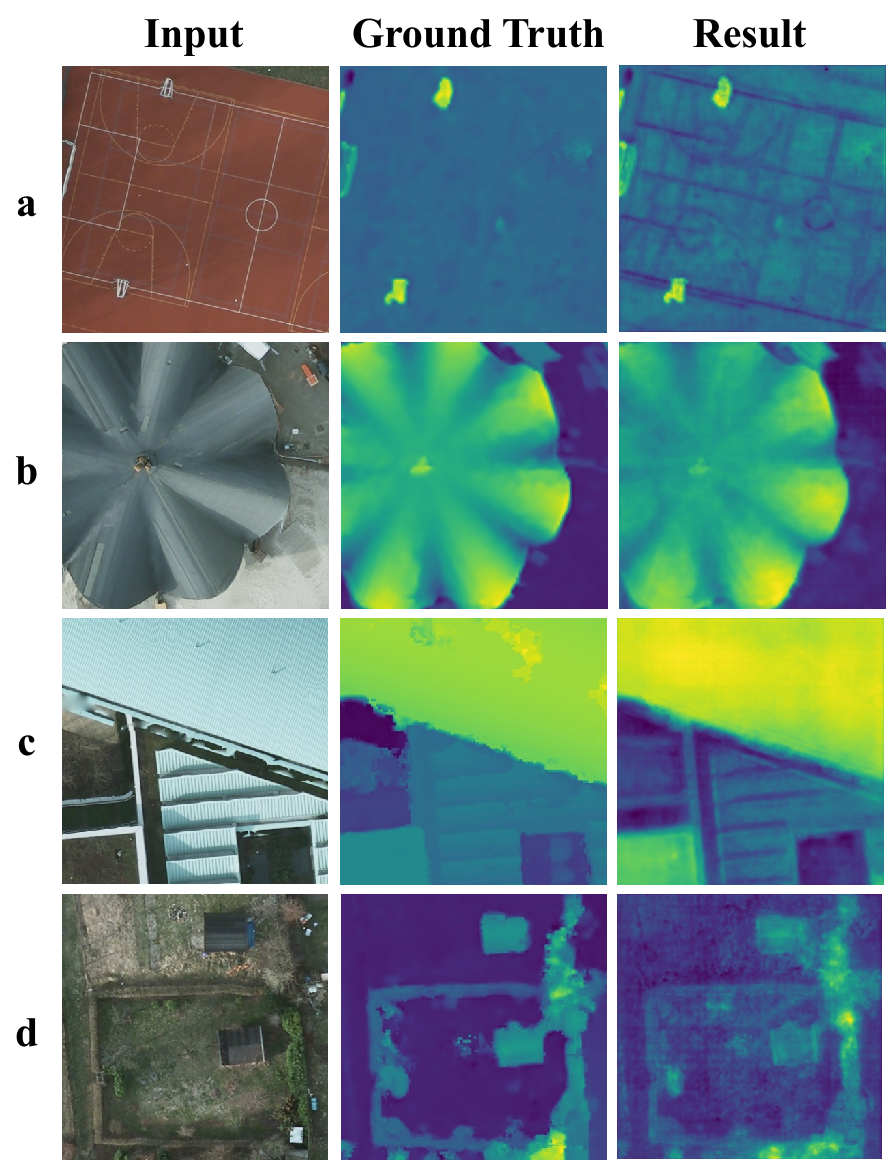} 
\caption{Visualization Results of Potsdam.} 
\label{fig9} 
\end{figure}\par
For the Potsdam dataset, a comparison of our proposed method, HeightFormer, was made with IMG2DSM, Amirkolaee et al. (\citeauthor{amirkolaee2019height}, \citeyear{amirkolaee2019height}), BAMTL(\citeauthor{9288901}, \citeyear{9288901}), DepthsFormer (\citeauthor{li2022depthformer}, \citeyear{li2022depthformer}), and BinsFormer (\citeauthor{li2022binsformer}, \citeyear{li2022binsformer}). The properties of the Potsdam dataset are presented in Table \ref{tbl5}, while the corresponding visualized results are illustrated in Fig.\ref{fig9}. The comparison revealed that HeightFormer, much like in the case of Vaihingen, accomplished state-of-the-art results in terms of Rel, RMSE (log), $\delta1$, $\delta2$, reducing the relative error to approximately 10\% (0.104) for the first time. Nevertheless, in terms of the metric $\delta3$, HeightFormer scored a bit lower than BinsFormer (0.997 vs 0.999). As inferred from the visualization, Fig. \ref{fig9} a and \ref{fig9} c demonstrate that, constrained by monocular information input, HeightFormer still presents deviation in the reconstruction of plane texture height, with a tendency to recover height-independent details such as white lines on the football field and building shadows . It is notable in Fig. \ref{fig9} d that, HeightFormer exhibits considerable reconstruction noise for complex plane instances, indicating a requirement for further improvements in the model's adaptability.\\\\
\section{Conclusion}
This study introduces Heightformer, a remote sensing monocular height estimation technique employing multilevel interaction and image-adaptive classification-regression. Our proposed Multilevel Interaction Backbone (MIB) capitalizes on attention mechanisms and the convolutional structure across varying interactive scales, extracting multi-scale information concurrently. This approach mitigates the pervasive issue of instance-level height bias within purely convolutional architectures. Additionally, the proposed Image-adaptive Classification-regression Height Generator (ICG) module reduces model convergence complexity and mitigates edge reconstruction ambiguity arising from fixed height value divisions. We assessed the effectiveness of our method using the ISPRS Vaihingen and Potsdam datasets, achieving relative errors of 0.185 and 0.104, respectively. Given that single-lens height estimation is inherently ill-posed, our HeightFormer model is compact and employs linear superimposition for height value generation, leading to the presence of non-height-related textures in the generated results (like Fig. \ref{fig9}). Moving forward, we intend to refine the Height Regression component of ICG, mitigate the instability induced by unlearnable dot product operations and bolster the model's generalization capacity.

\section*{Declaration of competing interest}
The authors declare that they have no known competing financial interests or personal relationships that could have appeared to influence the work reported in this paper.

\section*{Acknowledgments}
This work was supported by the Key Laboratory fund of Chinese Academy of Sciences under Grant CXJJ-23S032 and CXJJ-22S032.

\bibliographystyle{cas-model2-names}
\bibliography{refs.bib}
\end{document}